\documentclass[runningheads]{llncs}

\usepackage{eccv}

\usepackage{eccvabbrv}

\usepackage{graphicx}
\usepackage{booktabs}

\usepackage[accsupp]{axessibility}  

\usepackage{hyperref}

\usepackage{orcidlink}

\usepackage{multirow}

\begin{document}

\title{One Demonstration Is Enough for Real-World Robotic Reinforcement Learning}

\titlerunning{AutoSERL}

\author{Yuwan Liu\inst{1, 2, 3}\textsuperscript{*} \and
Hongze Yu\inst{3}\textsuperscript{*} \and
Song Liu\inst{3} \and
Yuhan Wang\inst{3} \and
Junge Zhang\inst{1, 4}\textsuperscript{\textdagger} \and
Yaodong Yang\inst{5}\textsuperscript{\textdagger} \and
Yuanpei Chen\inst{3} \and
Ceyao Zhang\inst{3, 5}\textsuperscript{\textdagger}}

\authorrunning{Y. Liu et al.}

\institute{
National Key Laboratory of Cognition and Decision Intelligence for Complex Systems, Institution of Automation, Chinese Academy of Sciences, Beijing, China \and Beijing Academy of Artificial Intelligence, Beijing, China \and PKU-PsiBot Joint Lab, Beijing, China \and School of Artificial Intelligence, University of Chinese Academy of Sciences, Beijing, China \and Institute for Artificial Intelligence, Peking University, Beijing, China}

\maketitle
\renewcommand{\thefootnote}{*}
\footnotetext[1]{Equal contribution.}
\renewcommand{\thefootnote}{\textdagger}
\footnotetext[2]{Corresponding authors. Email:  jgzhang@nlpr.ia.ac.cn, yaodong.yang@pku.edu.cn, ceyaozhang@pku.edu.cn}
\renewcommand{\thefootnote}{\arabic{footnote}} 

\begin{abstract}
Learning effective robot control policies on physical hardware is challenging due to costly data collection and the difficulty of reward specification. Prior work has incorporated demonstrations into reinforcement learning (RL), yet existing approaches either require large numbers of demonstrations or depend on continuous human intervention during training. To address these limitations, we present \textit{AutoSERL}, a framework that leverages a single demonstration to fully automate the intervention process in real-world robot RL. 
The framework includes three complementary mechanisms to accomplish certain tasks: a \textit{sliding window intervention} mechanism that continuously guides exploration to prevent local optima and unsafe deviations, a \textit{safety recovery mechanism} that detects and corrects failure states via predefined trajectory recovery points, and an \textit{intervention termination} criterion that automatically disables guidance once the policy can independently complete the task, preserving its exploration advantage.
We evaluate AutoSERL on six contact-intensive manipulation tasks across two robot platforms, spanning insertion, hanging, and hinge-based tasks. AutoSERL consistently outperforms SERL initialized with 20 demonstrations, behavior cloning, and MILES — a dedicated one-shot imitation learning baseline — across all tasks while matching HIL-SERL, achieves 100\% success rate on insertion tasks, and demonstrates improved robustness to positional variations, all from a single demonstration. Code and videos are available on our project website: https://autoserl.github.io/.
\keywords{Robotic Reinforcement Learning \and One-shot Learning \and Sample Efficiency}
\end{abstract}

\begin{figure}[t]
    \centering
    \includegraphics[width=\linewidth]{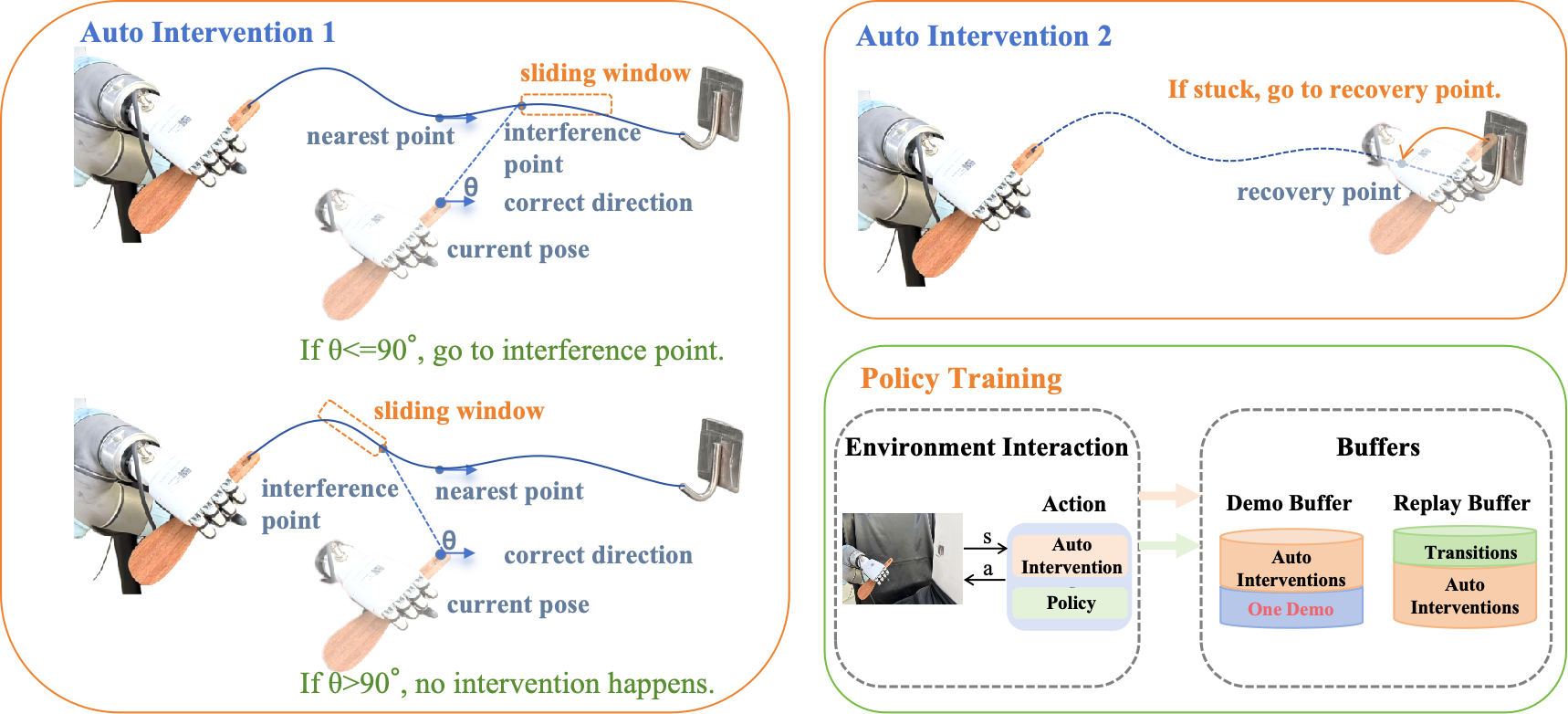}
    \caption{Overview of AutoSERL. \textbf{Auto Intervention 1} (Sliding Window 
    Intervention): the robot is guided to the nearest point within the sliding window 
    only when the angle $\theta$ between the trajectory's forward direction and the 
    vector to that point satisfies $\theta \leq 90^\circ$, preventing the robot from 
    being pulled back to already-visited positions. \textbf{Auto Intervention 2} 
    (Safety Recovery Mechanism): when the robot is stuck, it is guided to the recovery point and the demonstration segment is replayed to restore progress. 
    \textbf{Policy Training}: intervention-guided transitions and the single demonstration are stored in the Demo Buffer and Replay Buffer for policy training.}
    \label{fig:overview}
\end{figure}

\section{Introduction}\label{sec:intro}
Reinforcement learning (RL) has demonstrated remarkable success in simulated environments~\cite{mnih2013playing,silver2016mastering}, yet its deployment on physical robotic systems remains fundamentally constrained by two persistent challenges. First, real-world exploration carries significant inherent safety risks: unlike simulation where failures are costless, uncontrolled robot 
motions can cause hardware damage or environmental collisions~\cite{dulac2019challenges}. Second, the sample inefficiency of model-free RL means that agents often fail to encounter the rare and sparse reward signals necessary for policy convergence, particularly in contact-intensive tasks where successful 
states occupy only a small fraction of the state space.

To address these challenges, recent work has proposed combining demonstrations 
with RL to accelerate real-world learning~\cite{vecerik2017leveraging,rajeswaran2017learning,nair2018overcoming}. 
\cite{luo2024serl} introduces a sample-efficient framework that mixes demonstration data with online experience, significantly improving training stability on physical hardware. Building upon this, \cite{luo2025precise} incorporates human-in-the-loop intervention during training, where a human operator monitors the robot in real time and teleoperates it out of deadlocks or unsafe states. The corrected trajectories are added to the replay buffer, providing high-quality guidance that helps overcome sparse rewards and prevents 
unsafe exploration. This paradigm has proven effective for learning precise manipulation policies directly on real robots.

However reliance on continuous human intervention introduces a critical scalability bottleneck. Human operators are subject to cognitive fatigue, inconsistent response times, and high labor costs per training hour. As tasks grow more complex and training durations extend, the requirement for constant human presence becomes logistically impractical. This raises a fundamental question: \textit{can the benefits of human-in-the-loop training be preserved while eliminating the need for continuous human intervention?}

In this paper, we present \textit{AutoSERL}, a framework that replaces human intervention with automated intervention mechanisms derived from a \textbf{single demonstration}. 
The framework consists of three complementary components to accomplish certain tasks. A \textit{sliding window intervention} mechanism continuously monitors the 
agent's end-effector pose relative to a window sliding along the demonstration trajectory, detecting deviations caused by local optima, Q-value overestimation, or environmental obstacles, and redirecting the robot accordingly. A \textit{safety recovery mechanism} detects stagnation near interaction objects and triggers a replay-based recovery by re-executing a predefined segment of the demonstration trajectory, restoring the robot to a productive state without human intervention. An \textit{intervention termination} criterion monitors intervention frequency during each episode and automatically disables all guidance once the policy demonstrates sufficient autonomy, allowing the agent to fully leverage RL's exploratory advantage without unnecessary constraint.

We evaluate AutoSERL on six contact-intensive manipulation tasks spanning 
insertion, hanging, and hinge-based categories across two robot platforms. 
AutoSERL consistently outperforms \cite{luo2024serl} initialized with 20 
demonstrations, behavior cloning (BC), and MILES~\cite{papagiannis2024miles} --- a dedicated one-shot imitation learning method --- across all tasks, achieves 100\% success rate on insertion tasks, while achieving performance comparable to HIL-SERL\cite{luo2025precise}, and maintains strong performance under positional variations, all from a single demonstration. Our results demonstrate that one demonstration, when 
carefully structured into an automated intervention system, is sufficient to match and exceed the effectiveness of both multi-demonstration RL and human-supervised training.

The main contributions of this paper are as follows: 
1) we propose AutoSERL, a framework that automates real-world robot RL using only a single demonstration, eliminating the need for continuous human intervention; 
2) we design three complementary mechanisms --- sliding window intervention, safety recovery, and intervention termination --- that together form a closed-loop automated guidance system derived entirely from one demonstration trajectory; 3) we conduct extensive real-world experiments on six contact-intensive manipulation tasks across two robot platforms, demonstrating that AutoSERL consistently outperforms multi-demonstration RL, behavior cloning, and one-shot imitation learning baselines while matching HIL-SERL.

\section{Related Works}

\subsection{Human-in-the-Loop RL}
A prominent strategy to address the challenges of real-world exploration and sparse reward signals is to integrate human expertise directly into the training loop through real-time interventions. Paradigms such as Interactive Learning\cite{chen2025conrftreinforcedfinetuningmethod,liu2023robotlearningjobhumanintheloop,mandlekar2020humanintheloopimitationlearningusing,wu2025robocopilothumanintheloopinteractiveimitation} and DAgger-style imitation\cite{xu2025compliantresidualdaggerimproving,hoque2021thriftydaggerbudgetawarenoveltyrisk,kelly2019hgdaggerinteractiveimitationlearning,ross2011reductionimitationlearningstructured} allow human supervisors to provide corrective feedback or take over control when the agent enters a hazardous or non-productive state. By injecting this expert-induced bias, these methods transform random exploration into a structured data collection process, significantly accelerating policy convergence on physical hardware.
However, these methods suffer from a "human bottleneck." The need for constant vigilance creates a scalability paradox: while humans ensure safety, their presence restricts training duration and scale due to cognitive fatigue, latency, and high costs. Instead, AutoSERL achieves the benefits of corrective supervision without the need for a human in the loop.

\subsection{Safety and Exploration in Real-World RL}
An established approach to ensuring safety during real-world exploration is to employ Constrained Markov Decision Processes (CMDPs) or protective safety shields that filter out hazardous actions based on predefined constraints\cite{hu2025robottrainsrobotautomatic, thananjeyan2021recoveryrlsafereinforcement,li2020robustmodelpredictiveshielding,6225136, alshiekh2017safereinforcementlearningshielding, fisac2018generalsafetyframeworklearningbased}. By enforcing strict operational boundaries, these methods effectively safeguard hardware against catastrophic collisions during the learning process. However, such systems often require complex multi-robot coordination and specific hardware setups. Instead, our method achieves similar levels of autonomy and safety through purely rule-based geometric heuristics, requiring no additional robotic "teacher" or supervision. 

\subsection{Learning from Demonstration Replay}
A highly efficient approach to accelerating policy acquisition is to utilize replay-based imitation learning, which estimates the robot’s pose relative to objects of interest and re-executes demonstrated action sequences\cite{johns2021coarsetofineimitationlearningrobot, valassakis2022demonstrateonceimitateimmediately,dipalo2023effectivenessretrievalalignmentreplay,wen2022demonstrateoncecategorylevelmanipulation}. By leveraging existing expert trajectories, these methods provide a strong initial bias that bypasses the need for exhaustive random exploration in high-dimensional state spaces.
Instead, our approach utilizes a replay-based recovery protocol that leverages successful action segments from the demonstration trajectory. By re-executing these validated segments upon detecting stagnation, our framework provides a computationally efficient and robust guidance mechanism.

\section{Method}

\subsection{Preliminaries}
Reinforcement learning (RL) tasks can be modeled as a Markov Decision Process $\mathcal{M} = \left(\mathcal{S}, \mathcal{A}, \rho, P, r, \gamma\right)$, where $\mathcal{S}$ is the state space, $\mathcal{A}$ is the action space, $\rho(s_0)$ is the initial state distribution, $P$ is the state transition probability, $r$ is the reward function, and $\gamma$ is the discount factor. The learning objective is to maximize the expected cumulative reward. Our method is built upon SERL~\cite{luo2024serl}.

\subsection{Automated Intervention Design}
While SERL significantly improves sample efficiency, training can still fail on tasks with high difficulty and sparse rewards. To address this, \cite{luo2025precise} proposed HIL-SERL, which extends SERL by introducing human interventions during training to correct the robot's actions, thereby alleviating the negative impact of sparse rewards and enabling faster and more stable learning. However, HIL-SERL requires continuous human intervention throughout training, resulting in high labor costs. To reduce this burden, we analyze the situations in which human intervention is required in HIL-SERL, and use these empirical observations as the basis for designing AutoSERL.

Taking tasks that involve interaction between a hand-held object and a single
target object as an example, we identify three principal situations in which
intervention is needed. The first situation occurs when the training process
falls into a local optimum, where the magnitude of the output actions gradually
decreases and the robot arm keeps oscillating around a certain position. The
second situation occurs when the Q-value is incorrectly estimated during
training, causing the policy to continuously output actions that move the robot
away from the target object, which significantly reduces learning efficiency.
The third situation occurs when reinforcement learning is exploring normally,
but obstacles exist in the environment, potentially leading to
collisions with surrounding obstacles or causing the robot to get stuck. Beyond
these three situations, a residual risk exists: even when none of the above
conditions are met, the robot may still become physically stuck near the target object
and fail to make further progress. These four risks collectively define the
design requirements for AutoSERL.

\subsection{AutoSERL}
AutoSERL uses a single demonstration trajectory as guidance and addresses 
the identified risks through three complementary mechanisms, as illustrated in 
Fig.~\ref{fig:overview}. All tasks considered in this paper involve interaction between a hand-held 
object and a single target object. Before training begins, a demonstration trajectory 
is collected, and two recovery points are defined on it: 
$recover\ point_0$, a safe recovery target to which the robot can be guided 
when a safety risk occurs, and $recover\ point_1$, a trajectory point at which 
the robot is in stable contact with the interaction object. These two points 
can be identified by replaying the demonstration trajectory on the robot and 
serve as references across all three mechanisms. We also use an intervention termination threshold $th_1$ and an intervention start threshold $th_2$.

The first and second situations—local optimum convergence and erroneous 
Q-value estimation—along with the third situation involving environmental 
obstacles, are all effectively handled by the \textit{Sliding Window Intervention} 
mechanism, which actively guides the robot along the demonstration trajectory 
at every time step; since the trajectory is collected while maintaining a safe 
distance from obstacles, it naturally encodes obstacle avoidance. The residual 
stagnation risk is addressed by the \textit{Safety Recovery Mechanism}, a 
passive fallback that detects when the robot is stuck and restores it to a safe 
state. Finally, the \textit{Intervention Termination} criterion disables all 
interventions once the policy becomes capable of successfully completing the task 
independently, preventing continuous intervention from suppressing exploration.

\textbf{Sliding Window Intervention.} To leverage the demonstration trajectory for guiding the training process, we define a sliding window. The length of the sliding window is equal to the index difference between $recover$ $point_0$ and $recover$ $point_1$ on the demonstration trajectory. The window contains a continuous segment of trajectory indices.
At each time step, the system computes the distance between the current end-effector pose and those on the demonstration trajectory within the sliding window. The trajectory point with the minimum distance is selected and referred to as a potential interference point $pipoint$. When the minimum distance exceeds $th_2$, further judgment is required to determine whether intervention should be performed.

To avoid pulling the robot back to past trajectory positions, a directional check is conducted. $v_1$ denotes the tangent vector of the trajectory at the current $cpoint$ (the closest trajectory point to the end-effector), and $v_2$ denotes the vector from the current end-effector pose to the $pipoint$. The angle between these two vectors is computed. If the angle is greater than $90^\circ$, the $pipoint$ is considered to lie on a past segment of the trajectory, and the intervention is discarded. Otherwise, motion planning is used to move the robot to the $pipoint$ until the distance between the end-effector and the $pipoint$ is less than $th_1$.

At initialization, the end point of the sliding window is located at the first point of the demonstration trajectory. When no intervention occurs, the sliding window moves forward along the demo trajectory by one step at each time step. If the starting point of the sliding window reaches the end of the trajectory, the portion of the window that has reached the trajectory end will remain fixed, while the remaining portion that has not yet reached the end will continue to move forward. Therefore, using the sliding window for intervention can effectively prevent two situations: (1) the robot oscillating around a fixed position due to convergence to a local optimum; and (2) the policy continuously outputting actions that move the robot away from the target object due to inaccurate Q-value estimation. Moreover, since the demonstration trajectory is collected while maintaining a safe distance from environmental obstacles, this intervention strategy guides the robot toward the demonstration trajectory, thereby preventing collisions with surrounding obstacles.

\textbf{Safety Recovery Mechanism.} To ensure that the system can promptly recover to a controllable and safe state when safety risks occur during training, we define two recovery points on the demonstration trajectory: $recover$ $point_0$ and $recover$ $point_1$. $recover$ $point_0$ is defined as a safe recovery target. When a safety risk occurs, motion planning is used to guide the robot from the current state to $recover$ $point_0$. $recover$ $point_1$ is defined as a trajectory point where the robot is in stable contact with the interaction object. These two recovery points can be identified by replaying the demonstration trajectory on the robot.
During training, at each time step, the Euclidean distance between the current end-effector pose and all end-effector poses along the entire demonstration trajectory is computed. The trajectory point with the minimum distance is selected and referred to as the minimum point of the full trajectory $cpoint$. The system also maintains the minimum points over the most recent $l_{stag}$ time steps. If the distance between the current minimum point $cpoint_i$ and that from $l_{stag}$ time steps ago $cpoint_{i-l_{stag}}$ is smaller than the threshold $th_1$, and $cpoint_i$ lies between $recover$ $point_0$ and $recover$ $point_1$ on the demonstration trajectory, the robot is considered unable to properly interact with the object.
In this case, the system first uses motion planning to move the robot to $recover$ $point_0$, and then replays the demonstration trajectory segment between $recover$ $point_0$ and $recover$ $point_1$ to complete the recovery process.

\textbf{Intervention Termination.} Since the demonstration trajectory is not necessarily optimal, continuously applying intervention throughout training may reduce the policy’s exploration capability and limit the advantages of reinforcement learning. Therefore, in experiments without initial state randomization, if the total number of interventions in an episode is less than $l_{term}$ and the episode successfully completes the task, all subsequent interventions are disabled.

In all tasks considered in this paper, the threshold $th_1$ is set to 0.005 m, the threshold $th_2$ is set to 0.02 m, $l_{stag}$ is set to 20, and $l_{term}$ is set to 10.

\section{Experiments}

We conduct real-world experiments to evaluate the effectiveness of AutoSERL, aiming to address the following questions:
(1) Does AutoSERL achieve higher training efficiency than existing baselines?
(2) How robust is AutoSERL across different seeds and under positional variations?
(3) How do different heuristic hyperparameter settings affect the performance of AutoSERL?
(4) What are the respective contributions of the individual components of AutoSERL to its overall performance?
(5) Does AutoSERL merely imitate the demonstration trajectory, or can it learn to optimize beyond the demonstrated behavior?

\begin{figure}[ht]
    \centering
    \includegraphics[width=\linewidth]{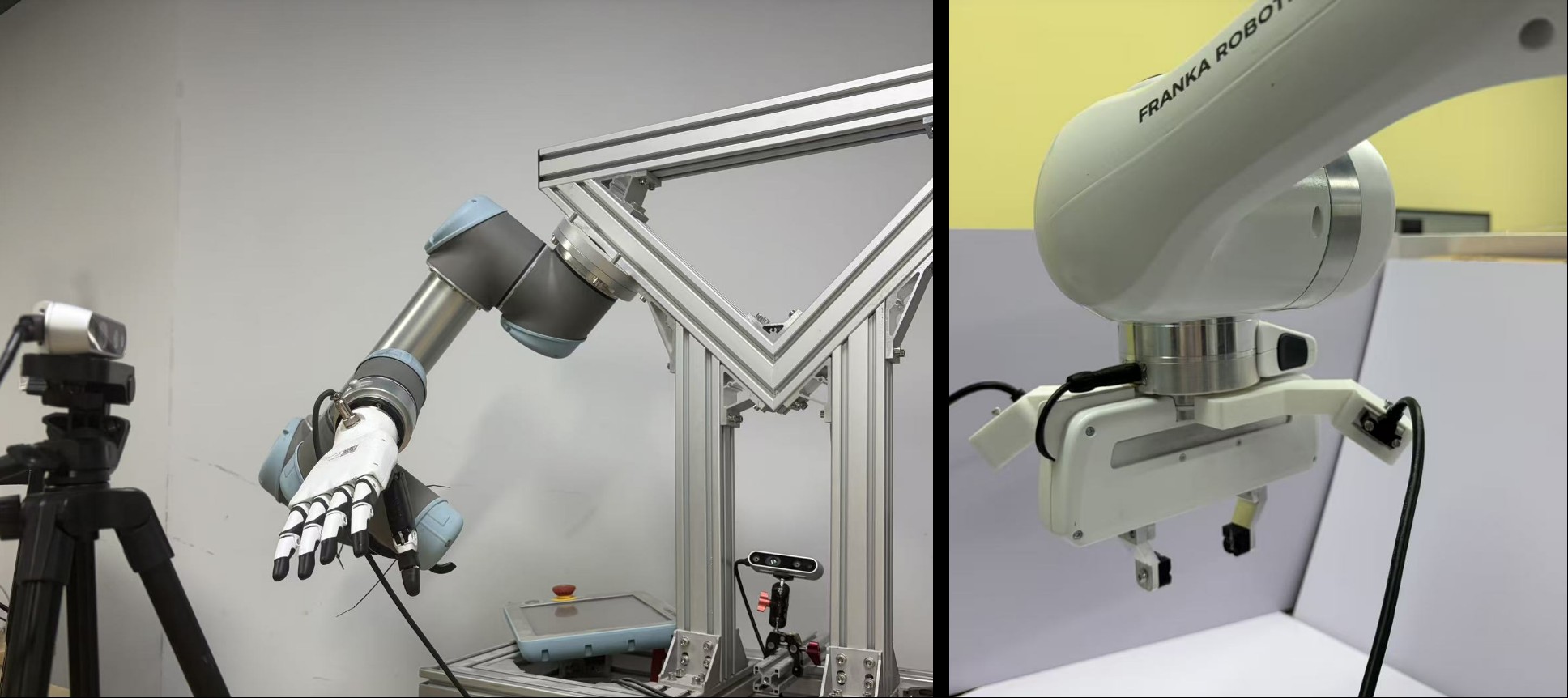}
    \caption{Experimental setup. Left: the setup for the hanging and hinge-based tasks, consisting of a UR5 robot, an Inspire dexterous hand and two Intel RealSense D435 cameras. Right: the setup for the insertion tasks, consisting of a Franka robot and two wrist-mounted Intel RealSense D405 cameras.}
    \label{fig:settings}
\end{figure}

\subsection{Experimental Setup}

\textbf{Hardware and Protocol.}
The insertion tasks are conducted on a Franka robot arm with a parallel gripper 
and two wrist-mounted Intel RealSense D405 cameras. The hanging and hinge-based 
tasks are implemented on a UR5 robot equipped with an Inspire dexterous hand 
and two Intel RealSense D435 cameras. The overall setup of the two robotic platforms is shown in 
Fig.~\ref{fig:settings}. Across all tasks, the observation consists of RGB images 
from the two cameras and robot proprioceptive states. For the Franka platform, 
proprioception includes end-effector poses, velocity, forces/torques, and 
gripper state. For the UR5 platform, proprioception includes end-effector poses 
and forces/torques. The action space is a 6D delta end-effector pose for all 
tasks. Training uses manually annotated binary sparse rewards. Episodes 
terminate upon task success or after 300 time steps. During evaluation, all 
automatic intervention mechanisms are disabled, and each task is evaluated over 
50 episodes.

\begin{figure}[t!]
    \centering
    \includegraphics[width=\linewidth]{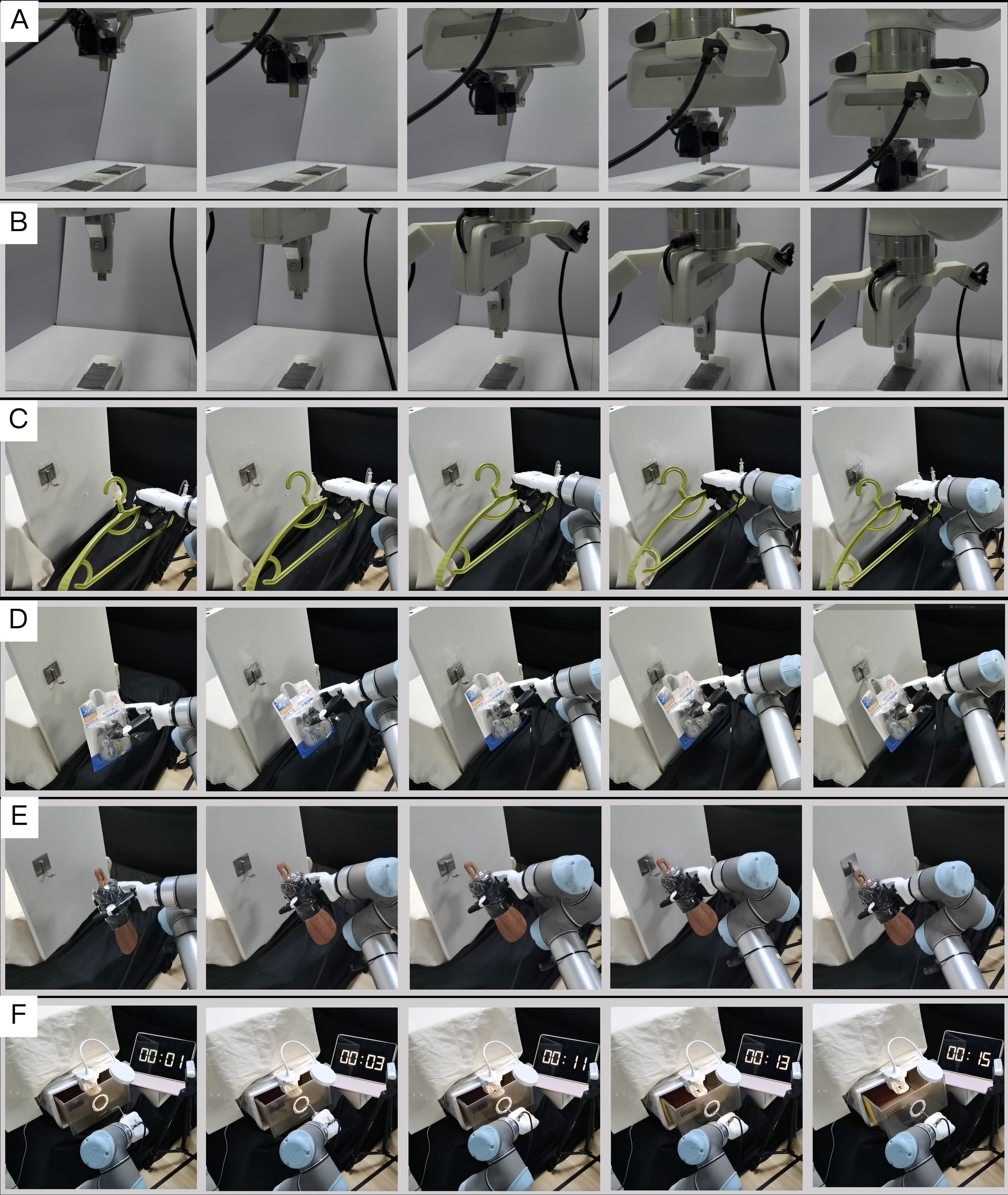}
    \caption{Overview of the experimental tasks: (A)Plug Insertion. (B)USB Insertion. (C)Hanger Suspension. (D)Correction Tape Suspension. (E)Spoon Suspension. (F)Drawer Opening.}
    \label{fig:all_task}
\end{figure}

\begin{figure}[th]
    \centering
    \includegraphics[width=.8\linewidth]{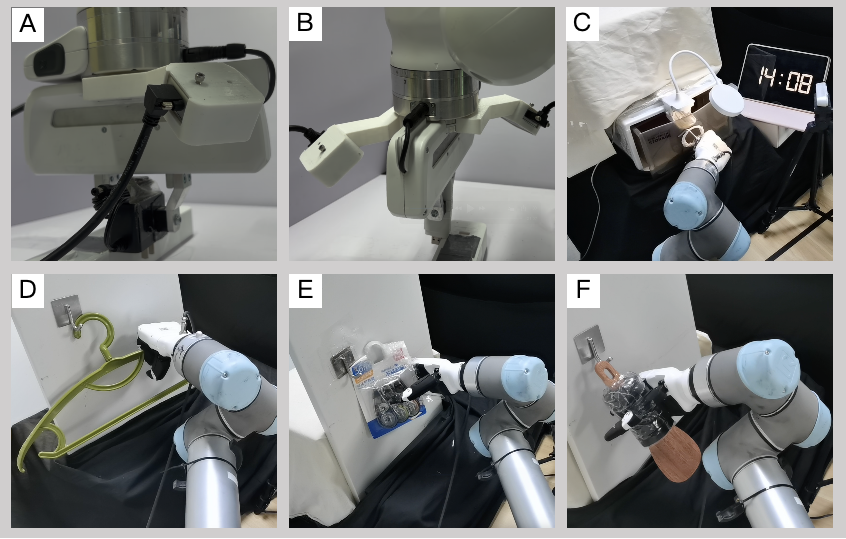}
    \caption{Overview of the stuck cases across different tasks: (A)Plug Insertion. (B)USB Insertion. (C)Drawer Opening. (D)Hanger Suspension. (E)Correction Tape Suspension. (F)Spoon Suspension.}
    \label{fig:kazhu}
\end{figure}

\textbf{Task Overview.}
We implement three categories of contact-intensive manipulation tasks, as 
illustrated in Fig.~\ref{fig:all_task}: insertion tasks (plug insertion and USB 
insertion), hanging tasks (hanging a correction tape, a hanger, and a spoon), 
and a hinge-based task (drawer pulling using a hook). These tasks are 
characterized by rich physical contact and low tolerance for execution errors, 
requiring high manipulation precision, accurate pose alignment, stable contact 
control, and strong robustness to disturbances.
If the above conditions are not satisfied, the robot may easily become stuck, as shown in Fig.~\ref{fig:kazhu}.

\begin{figure}[t!]
    \centering
    \includegraphics[width=\linewidth]{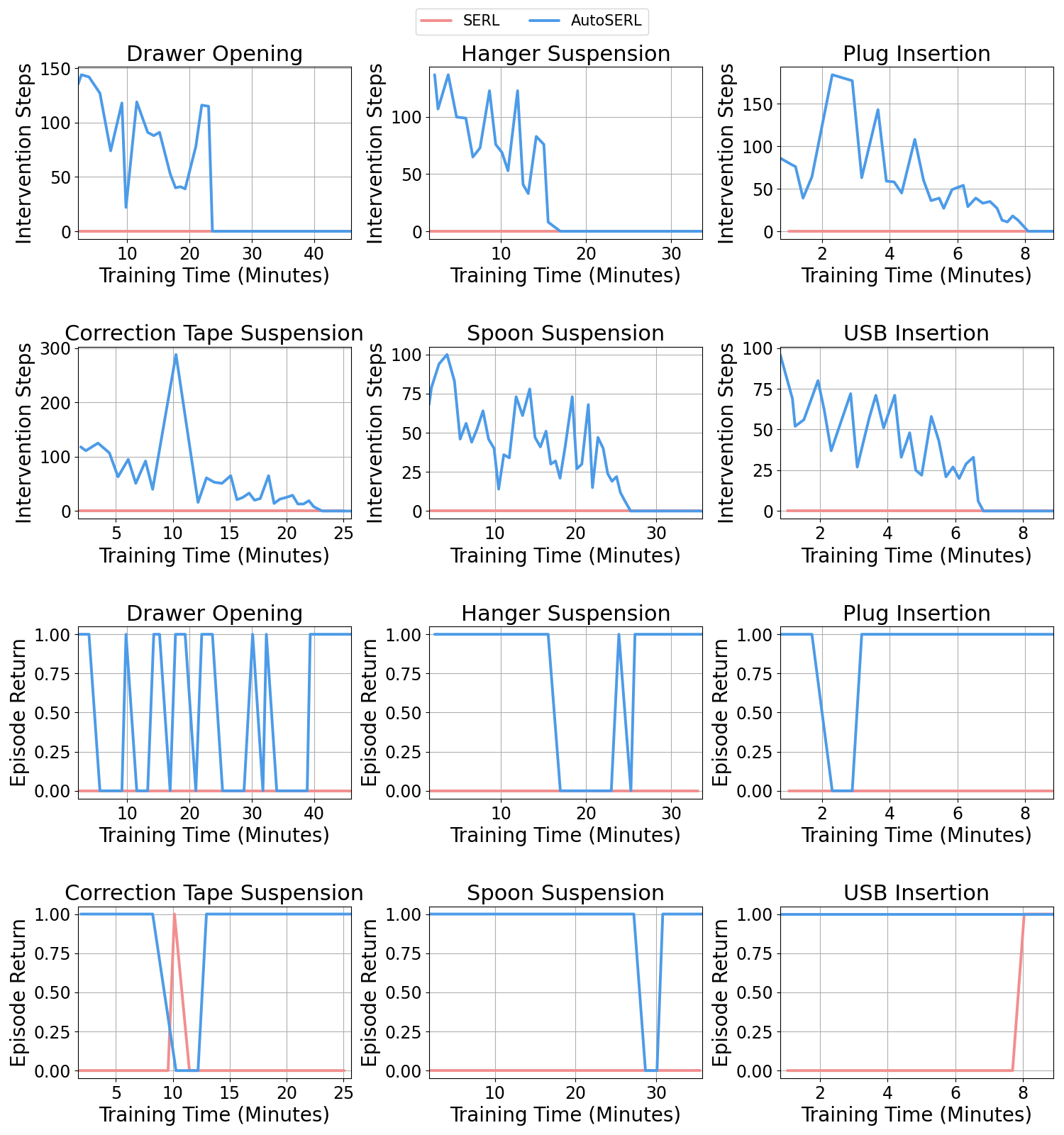}
    \caption{Training curves of time versus intervention steps and time versus episode return for each task under SERL and AutoSERL.}
    \label{training_figs}
\end{figure}

\subsection{Results and Analysis}

\textbf{Training Efficiency.}
We compare AutoSERL with SERL~\cite{luo2024serl} and HIL-SERL~\cite{luo2025precise} to evaluate training efficiency. In this set of experiments, no randomization is applied to any of the task scenarios. For each task, both SERL and HIL-SERL are initialized with 20 demonstration trajectories. AutoSERL and SERL are assessed by measuring their success rates under identical training durations.
As shown in Table~\ref{tab:serl_autoserl}, AutoSERL achieves higher success rates than SERL under the same training duration, demonstrating improved sample efficiency. Fig.~\ref{training_figs} presents the training curves for each task. The intervention step curve shows that the number of automatic interventions gradually decreases as training progresses, indicating that the learned policy increasingly approximates the demonstrated behavior. The episode return curve shows that AutoSERL can stably accomplish the tasks without intervention after the same training duration, whereas SERL fails to achieve consistent task success.
We compare AutoSERL with HIL-SERL by measuring the minimum training time required to reach a 100\% success rate. As shown in Table~\ref{tab:hilserl_comparison}, AutoSERL requires less or equal training time than HIL-SERL on five out of six tasks, indicating that the proposed automatic intervention mechanism provides guidance comparable to human interventions.

\textbf{Comparison with Baselines.}
To further validate the effectiveness of AutoSERL, we compare it with 
BC and MILES~\cite{papagiannis2024miles} in terms of 
final success rate. For BC, 1 demonstration is used for the hinge-based task, 
10 demonstrations for the hanging tasks, and 20 demonstrations for the 
insertion tasks. For MILES, we adopt the same data augmentation randomization 
range as reported in the original work during data collection, namely $\pm$4~cm in translation and 
$\pm$4$^\circ$ in rotation. For all methods, no task scenario randomization is introduced during evaluation. The results are summarized in 
Table~\ref{tab:baseline_comparison}. AutoSERL consistently outperforms both 
baselines across all tasks, demonstrating its superior performance in 
real-world manipulation scenarios.

\begin{table}[t]
\centering
\caption{Success rate of SERL and AutoSERL under the same training time budget.}
\label{tab:serl_autoserl}
\begin{tabular}{l|c|c|c}
\toprule
\multirow{2}{*}{Task} & \multirow{2}{*}{Training Time (min)} & \multicolumn{2}{c}{Success Rate} \\
\cmidrule(lr){3-4}
 &  & SERL & AutoSERL \\
\midrule
USB Insertion & 8  & 20/50 & 50/50 \\
Plug Insertion & 8  & 0/50  & 50/50 \\
Hanger Suspension & 33 & 0/50  & 50/50 \\
Correction Tape Suspension & 25 & 6/50  & 50/50 \\
Spoon Suspension & 35 & 0/50  & 50/50 \\
Drawer Opening & 45 & 0/50  & 50/50 \\
\bottomrule
\end{tabular}
\end{table}

\begin{table}[t]
\centering
\caption{Minimum training time required to achieve a 50/50 success rate for AutoSERL and HIL-SERL.}
\label{tab:hilserl_comparison}
\begin{tabular}{lcc}
\toprule
{Task} & \multicolumn{2}{c}{Training Time (min)} \\
\cmidrule(lr){2-3}
 & AutoSERL & HIL-SERL \\
\midrule
USB Insertion & 8 & 6 \\
Plug Insertion & 8 & 8 \\
Hanger Suspension & 33 & 48 \\
Correction Tape Suspension & 25 & 60 \\
Spoon Suspension & 35 & 70 \\
Drawer Opening & 45 & 45 \\
\bottomrule
\end{tabular}
\end{table}

\begin{table}[t!]
\centering
\caption{Success rate comparison between AutoSERL and baselines.}
\label{tab:baseline_comparison}
\begin{tabular}{lccc}
\toprule
Task & AutoSERL & BC & MILES \\
\midrule
USB Insertion & 50/50 & 5/50  & 0/50  \\
Plug Insertion & 50/50 & 2/50  & 33/50 \\
Correction Tape Suspension & 50/50 & 38/50 & 1/50  \\
Hanger Suspension & 50/50 & 37/50 & 42/50 \\
Spoon Suspension & 50/50 & 0/50  & 2/50  \\
Drawer Opening & 50/50 & 35/50 & 0/50  \\
\bottomrule
\end{tabular}
\end{table}

\begin{figure}[t!]
    \centering

    \begin{subfigure}[b]{0.24\linewidth}
        \centering
        \includegraphics[width=\linewidth]{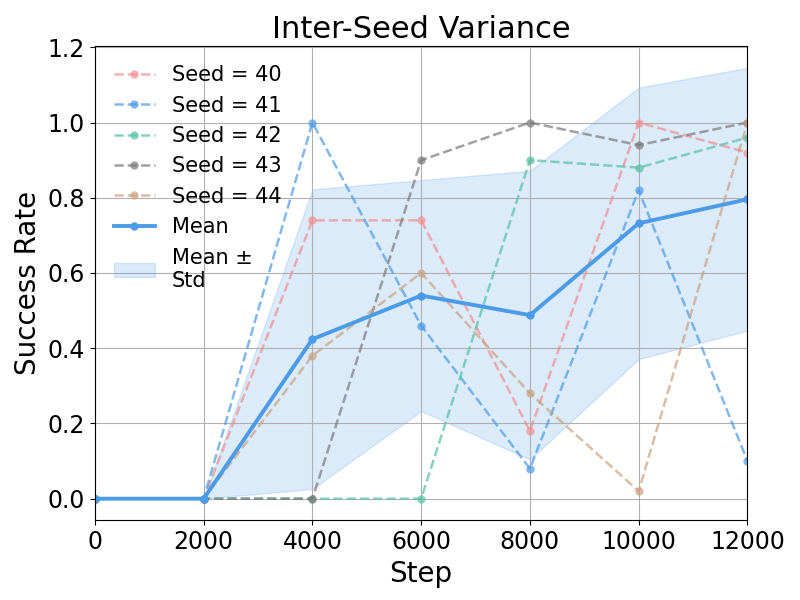}
        \caption{}
        \label{fig:inter-seed_variance}
    \end{subfigure}
    \hfill
    \begin{subfigure}[b]{0.24\linewidth}
        \centering
        \includegraphics[width=\linewidth]{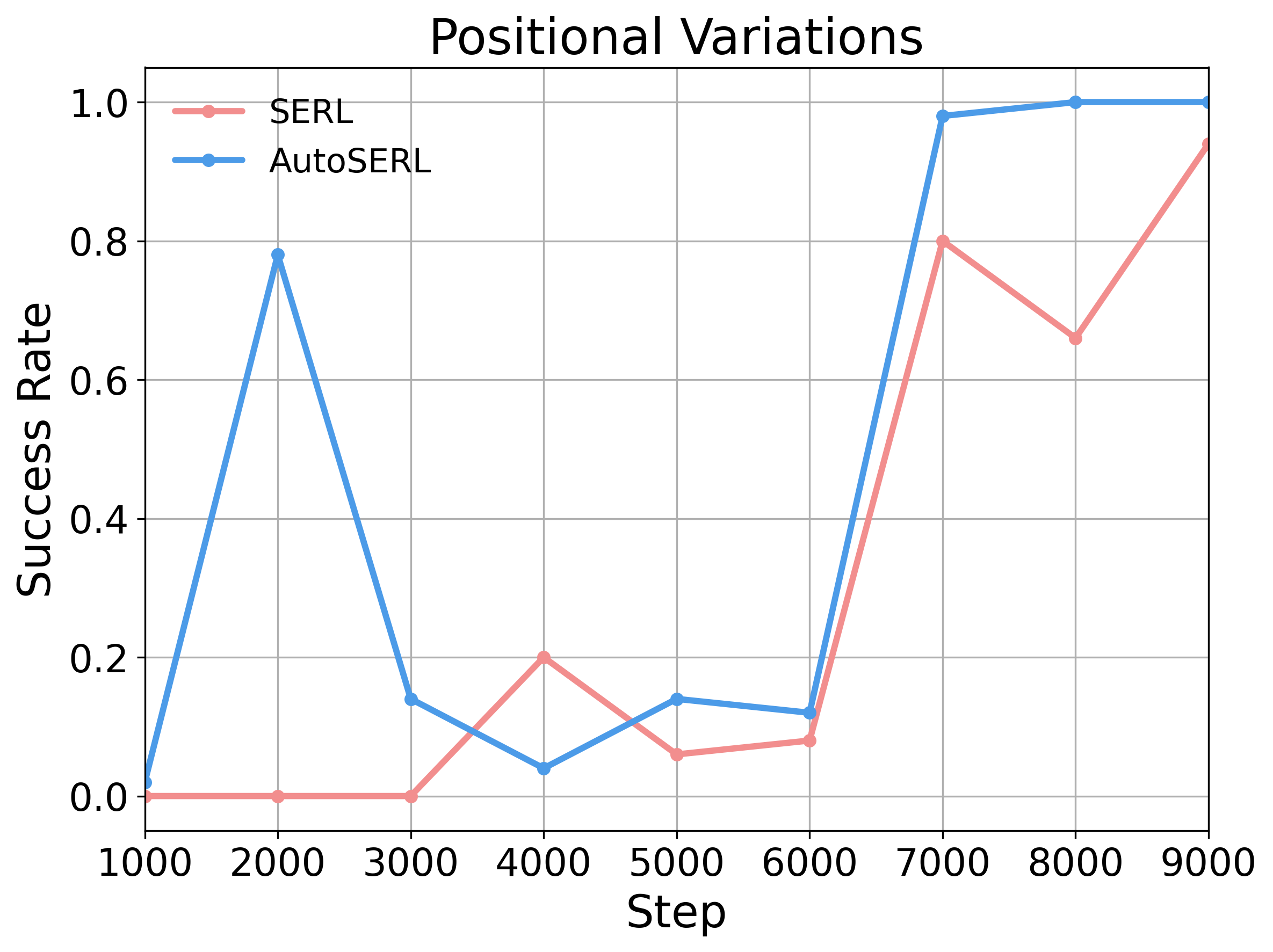}
        \caption{}
        \label{fig:positional_variations}
    \end{subfigure}
    \hfill
    \begin{subfigure}[b]{0.24\linewidth}
        \centering
        \includegraphics[width=\linewidth]{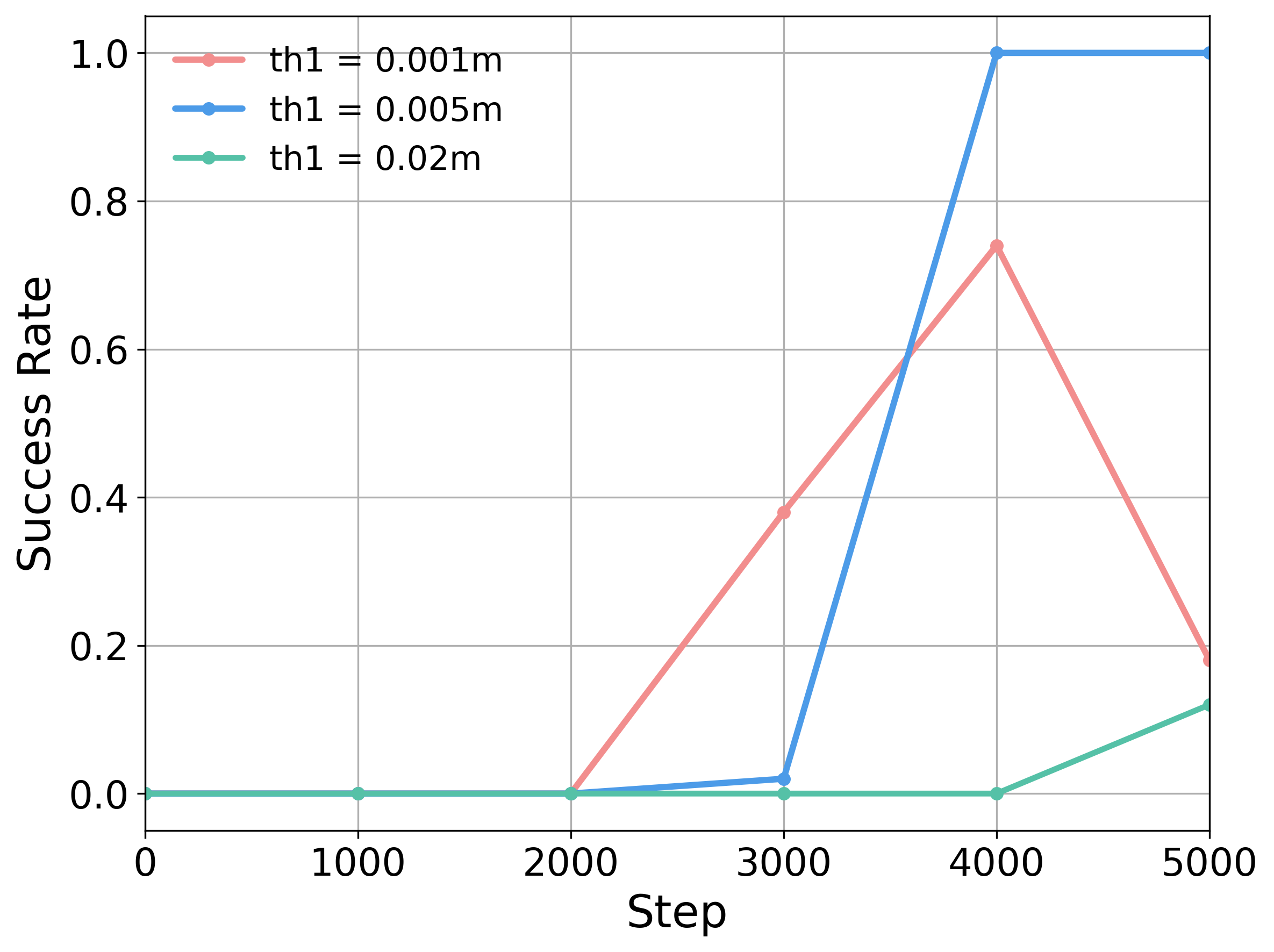}
        \caption{}
        \label{fig:params_0}
    \end{subfigure}
    \hfill
    \begin{subfigure}[b]{0.24\linewidth}
        \centering
        \includegraphics[width=\linewidth]{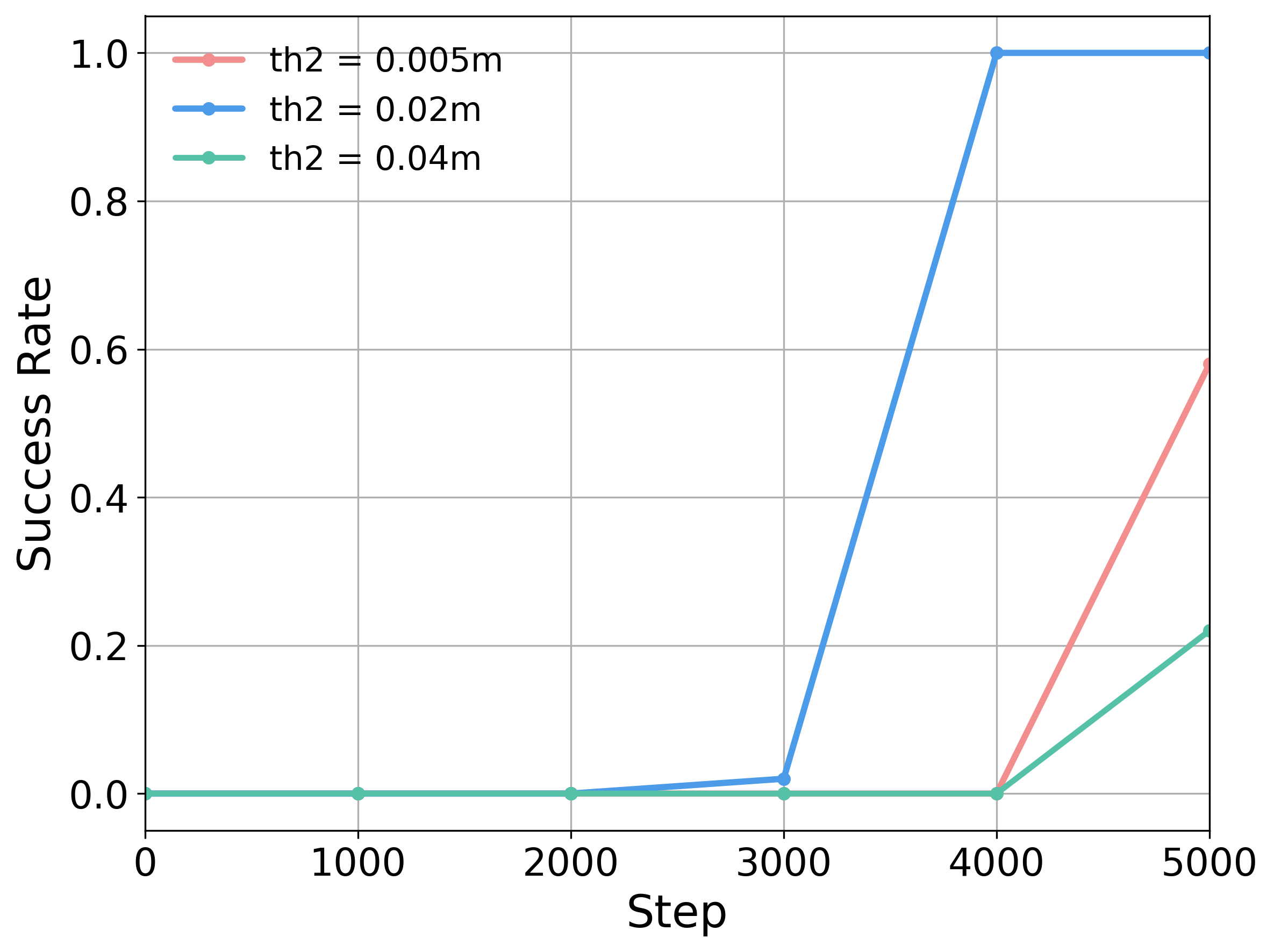}
        \caption{}
        \label{fig:params_1}
    \end{subfigure}
    \hfill
    \caption{Robustness and Heuristic Hyperparameter Analysis: (a) and (b) show the training curves for the plug insertion task across five random seeds and under positional variations. (c) and (d) present the training curves for the plug insertion task under different settings of hyperparameters $th_1$ and $th_2$, respectively.}
    \label{fig:robustness_params_total}
\end{figure}

\subsection{Robustness Analysis}
\textbf{Inter-seed variance.} We retrain the plug insertion task with five seeds (40–44). The success curves and summary statistics are shown in Fig.~\ref{fig:robustness_params_total}(a). Under all seeds, the method achieves 100\% or near 100\% success rates, indicating robustness to random initialization and low performance variance across different seeds.

\textbf{Positional variations.} In the plug insertion task, we randomize the initial plug position within a $\pm$3~cm range in the 
$x$--$y$ plane while keeping the socket position fixed to evaluate robustness to positional variations. For each episode, the intervention reference trajectory consists of the original demonstration trajectory concatenated with a motion-planned segment connecting the randomized initial point to the original starting point. Since 
positional variations alter the distribution of intervention triggers during 
training, all intervention mechanisms are kept active throughout the training 
phase of this experiment; evaluation follows the standard protocol with all 
interventions disabled. Under this setting, we compare SERL and AutoSERL using 
the same training configuration and report training-time versus success-rate 
curves. As illustrated in Fig.~\ref{fig:robustness_params_total}(b), 
AutoSERL achieves higher success rates and more stable convergence than SERL, 
demonstrating improved robustness to positional variations.

\subsection{Heuristic Hyperparameter Analysis}
We introduce several heuristic parameters and evaluate multiple values for each on the plug insertion task, with success-rate curves over time steps shown in Fig.~\ref{fig:robustness_params_total}(c)(d). As shown in the figure, we study the sensitivity of $th_1$ and $th_2$. For $th_1$, we test 0.001 m and 0.02 m against the default 0.005 m; for $th_2$, we test 0.005 m and 0.04 m against 0.02 m. In both cases, the alternatives degrade performance. For $th_1$, too small a value delays reaching the interference point, while too large a value weakens intervention effectiveness due to large post-intervention deviation. For $th_2$, too small a value causes overly frequent interventions, reducing exploration, while too large a value leads to insufficient interventions. Overall, both parameters degrade performance when set too small or too large, and perform best within a moderate range.
The stagnation window length $l_{stag}$ defines the number of steps required for the end-effector to move beyond $th_1$ and is set according to the task’s action scale.

\begin{figure}[t!]
    \centering

    \begin{subfigure}[b]{0.24\linewidth}
        \centering
        \includegraphics[width=\linewidth]{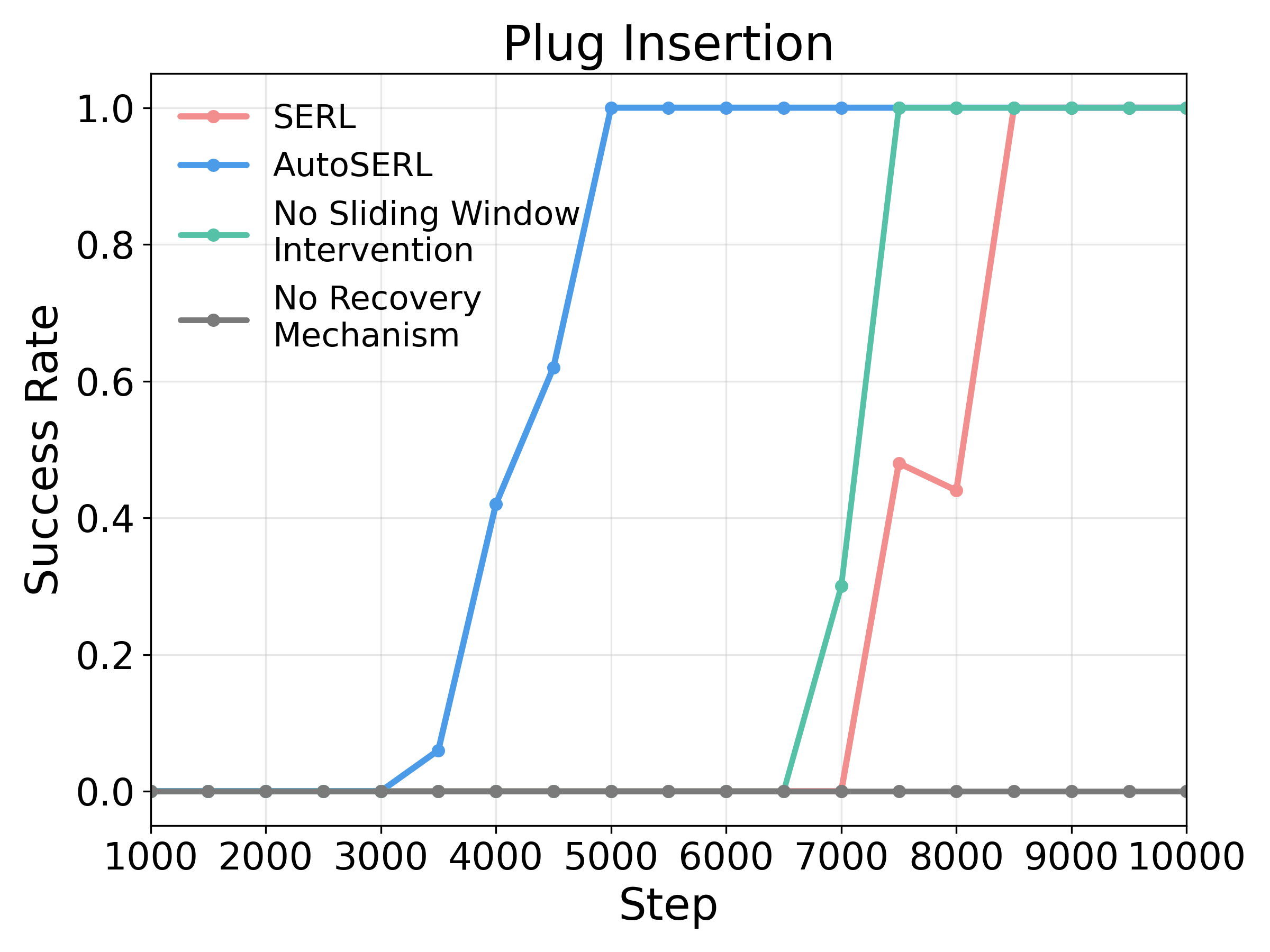}
        \caption{}
        \label{fig:ablation_plug}
    \end{subfigure}
    \hfill
    \begin{subfigure}[b]{0.24\linewidth}
        \centering
        \includegraphics[width=\linewidth]{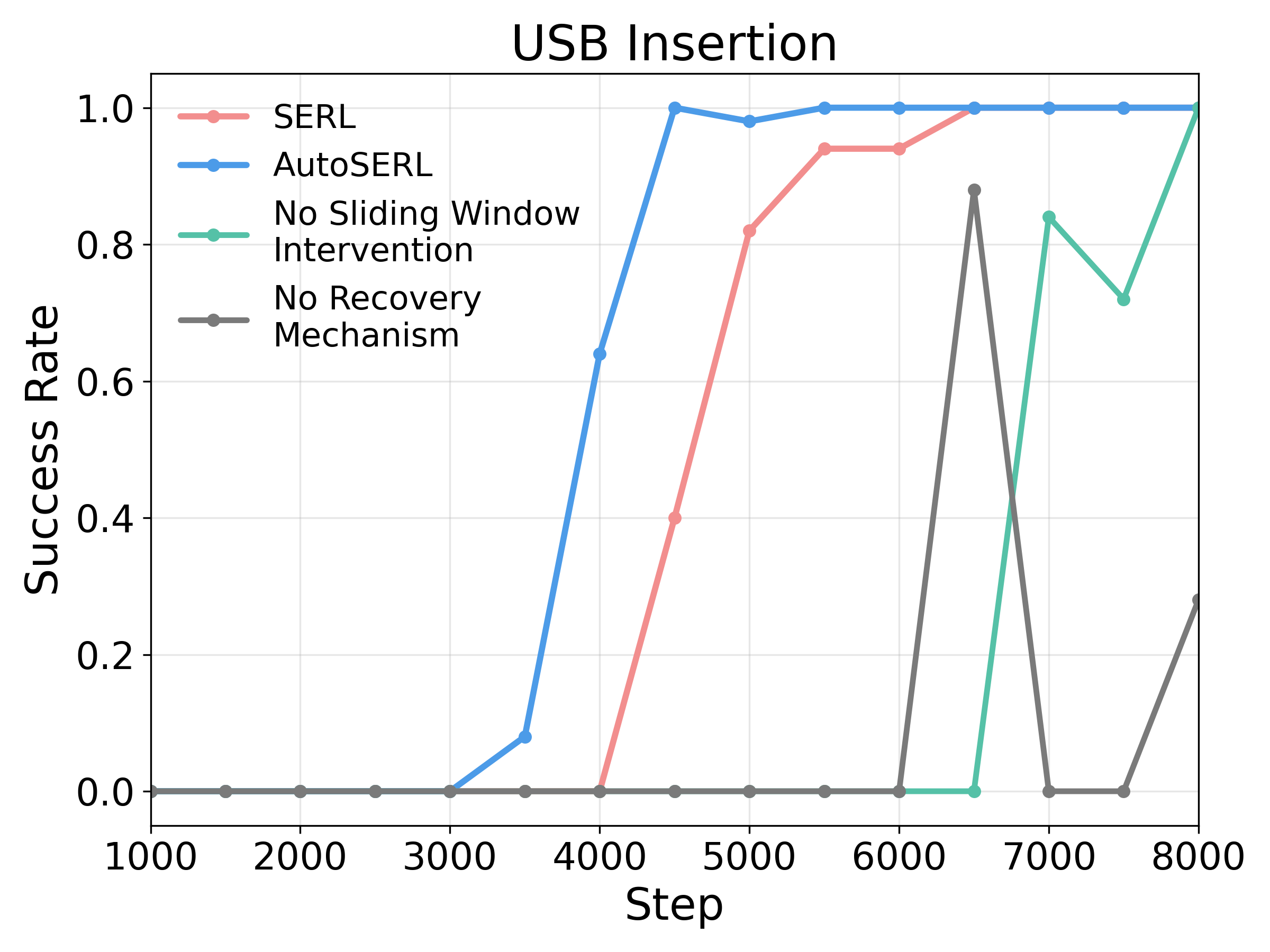}
        \caption{}
        \label{fig:ablation_usb}
    \end{subfigure}
    \hfill
    \begin{subfigure}[b]{0.24\linewidth}
        \centering
        \includegraphics[width=\linewidth]{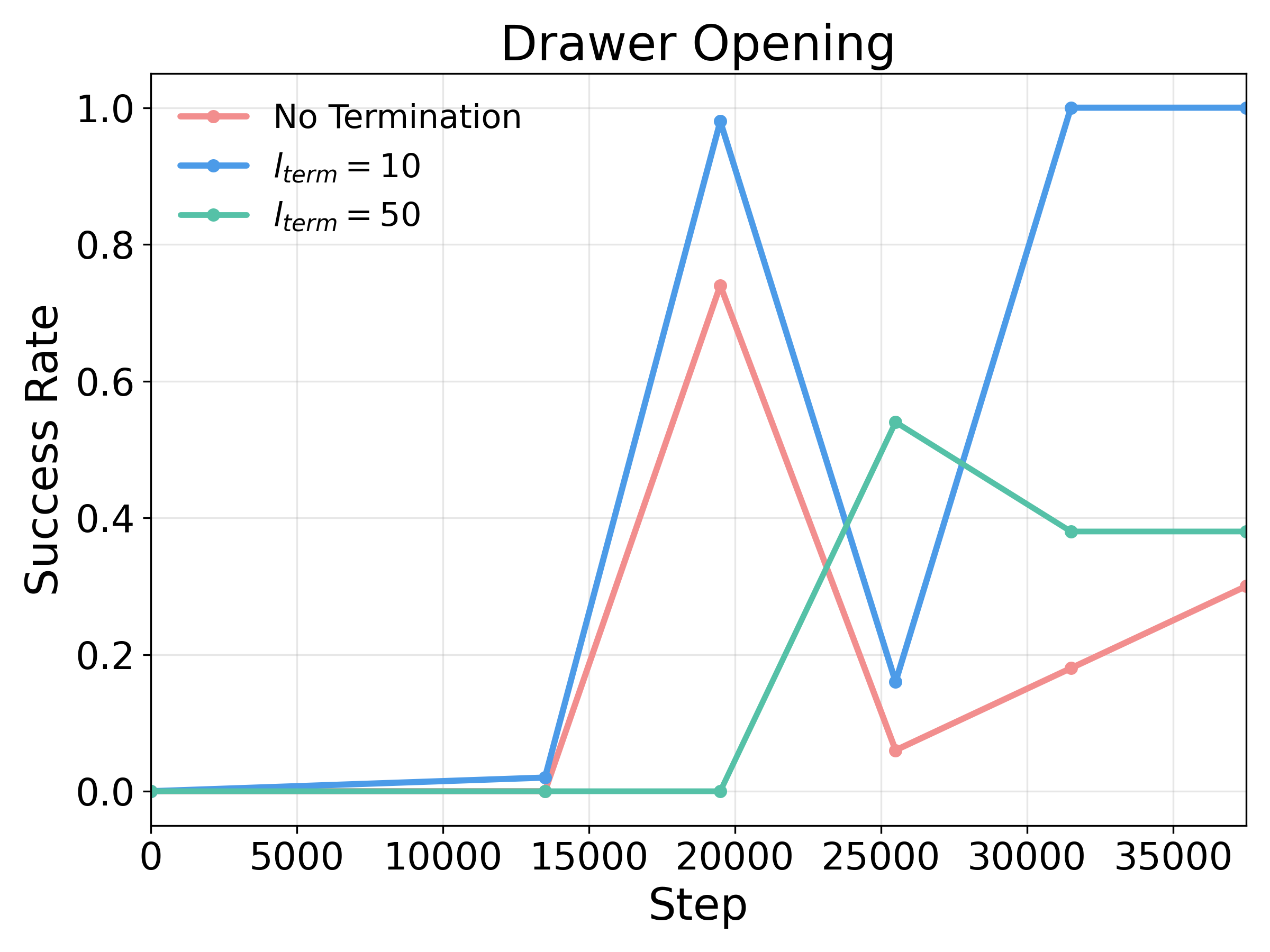}
        \caption{}
        \label{fig:ablation_drawer}
    \end{subfigure}
    \hfill
    \begin{subfigure}[b]{0.24\linewidth}
        \centering
        \includegraphics[width=\linewidth]{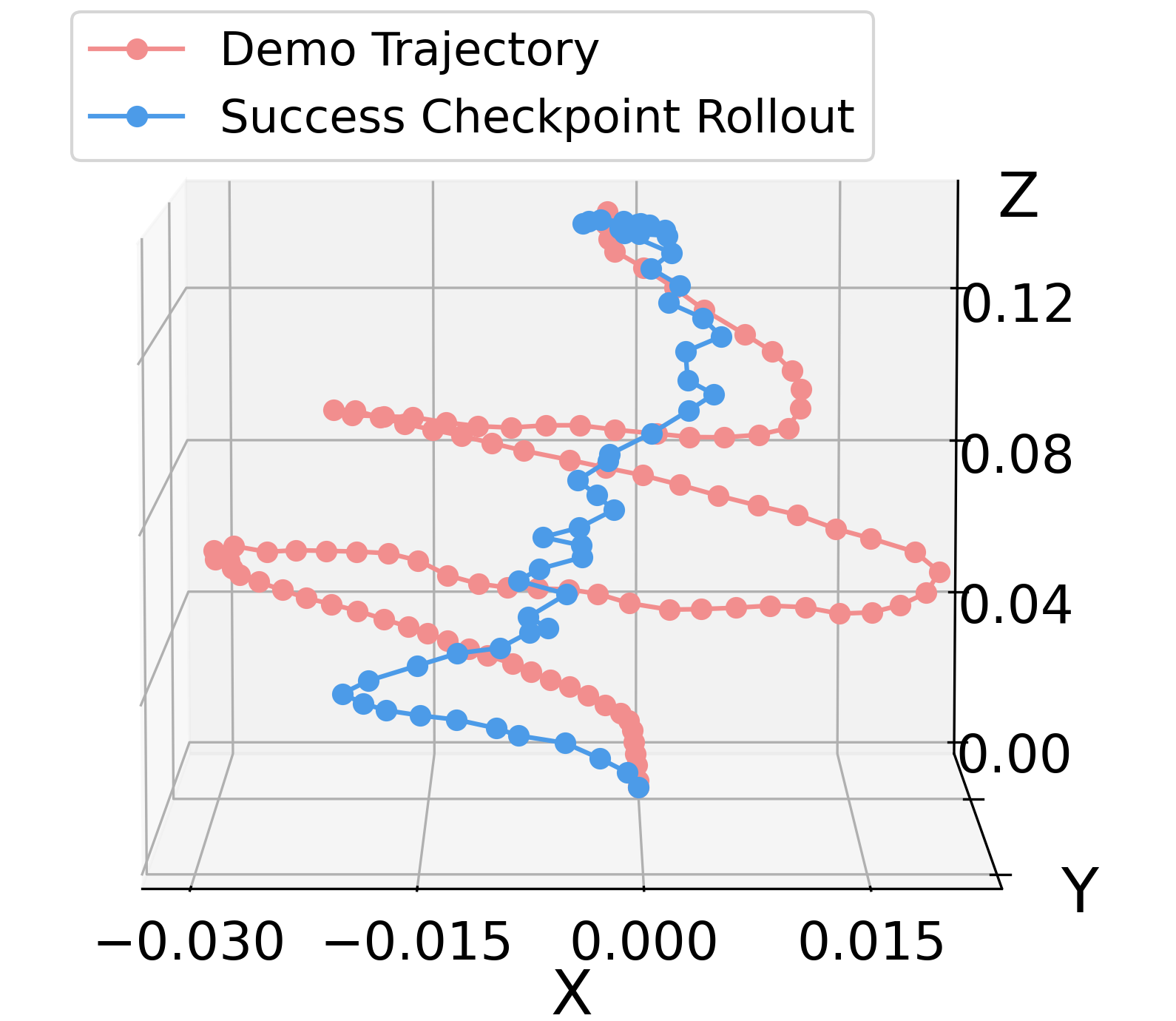}
        \caption{}
        \label{fig:traj_comparison_3d}
    \end{subfigure}
    \hfill
    
    \caption{
Ablation study and trajectory comparison: (a) and (b) results on the plug insertion and USB insertion tasks under the \textit{No sliding window intervention} and \textit{No recovery mechanism} settings, respectively. (c) results on the drawer opening task under the \textit{No intervention termination} setting. (d) results on 3D visualization of the demo trajectory and the policy rollout trajectory trained from it for the plug insertion task.}
    \label{fig:ablation_traj_comparation}
\end{figure}

\subsection{Ablation Studies}
To evaluate the contribution of each component in AutoSERL, we conduct three ablation studies: (1) \textit{No sliding window intervention}: the sliding window intervention mechanism is removed, and no interference points are used to guide the robot during training. 
(2) \textit{No recovery mechanism}: the safety recovery mechanism is removed, and when the robot encounters failure states, it must rely solely on its own exploration to recover. 
(3) \textit{No intervention termination}: the intervention termination criterion is removed, and the intervention continuously supervises the training process until the end.

For the \textit{No sliding window intervention} and \textit{No recovery mechanism} settings, we conduct experiments on the plug insertion and USB insertion tasks. The results are shown in Fig.~\ref{fig:ablation_traj_comparation}(a)(b). The full AutoSERL model achieves a 100\% success rate the earliest. Removing the sliding window intervention mechanism increases the number of training steps required to reach 100\% success. Removing only the recovery mechanism while retaining the sliding window intervention mechanism leads to worse performance than SERL, as the sliding window strategy may guide the robot into near-failure states. Without an explicit recovery mechanism, the policy must rely solely on sparse-reward exploration to escape these states, resulting in higher failure rates and less stable learning.

For the \textit{No intervention termination} setting, we study the mechanism on the drawer opening task under three configurations: no termination, $l_{term}$=10, and $l_{term}$=50. The corresponding success curves are shown in Fig.~\ref{fig:ablation_traj_comparation}(c). Without termination, performance peaks at approximately 19,500 steps but subsequently degrades, as continuous intervention leads to over-reliance on external corrections or frequent recovery triggers caused by misjudgments of stagnation due to small action magnitudes, thereby weakening autonomous learning. In contrast, introducing termination provides early-stage guidance with reduced exploration space, followed by a transition to autonomous reinforcement learning, improving overall performance. Moreover, $l_{term}$=10 outperforms $l_{term}$=50, indicating that overly large termination steps end intervention before sufficient guidance is obtained. Therefore, the mechanism is necessary, and its hyperparameters need to balance insufficient guidance and over-constrained learning.

\subsection{Policy and Demonstration Trajectory Comparison}
Taking the plug insertion task as an example, we collect a non-optimal demonstration trajectory of length 99 and train the policy using this trajectory. Rolling out the first checkpoint that achieves a 50/50 success rate yields a trajectory of length 54. The 3D visualization of both trajectories is shown in the Fig.~\ref{fig:ablation_traj_comparation}(d). This suggests that the policy goes beyond simple imitation and performs trajectory-level optimization over the demonstration.

\section{Discussions and Limitations}
Although AutoSERL shows strong performance in real-world manipulation tasks, several limitations remain. 
First, for tasks with highly diverse failure modes, AutoSERL’s recovery mechanism cannot recover from failures effectively during training because it relies on information from only a single trajectory. FARL~\cite{li2026failure} and UniIntervene~\cite{deng2026uniintervene} train their recovery policies using multiple trajectories. Moreover, UniIntervene points out that when the encountered failure situation falls outside the training distribution, the reliability of recovery can be significantly reduced. Therefore, a promising research direction is to leverage more data to train a more robust recovery policy, enabling automated real-world robot RL for a wider range of tasks.
Second, the current framework is limited to tasks with a 6D delta end-effector pose action space, and generalizing automated real-world robotic RL to higher-dimensional action spaces remains future work.

\section{Conclusion}
We propose AutoSERL, a real-world RL method that enables training through automatic intervention using only a single demonstration trajectory. AutoSERL incorporates sliding window intervention, safety recovery, and intervention termination to ensure safe and stable real-world reinforcement learning.
AutoSERL outperforms multi-demonstration RL, behavior cloning, and one-shot imitation learning baselines across six contact-intensive tasks spanning three task categories while matching HIL-SERL.
We hope this work will inspire future research on automated real-world robotic RL and enable its extension to tasks with more diverse failure modes and higher-dimensional action spaces.


\section*{Acknowledgements}
We gratefully acknowledge Borong Zhang for his valuable suggestions. We also thank Yu Li and Yishuai Cai for support in the initial setup of the Franka robotic arm system. This work is supported by the Chinese Academy of Sciences Project for Young Scientists in Basic Research (Grant No. YSBR-107) and the National Natural Science Foundation of China (Grant Nos. 62376013 and 62561160152).

%
%
\bibliographystyle{splncs04}
\bibliography{main}

\newpage
\section*{Appendix}
\appendix
\setcounter{figure}{0}
\setcounter{table}{0}

\renewcommand{\thefigure}{A.\arabic{figure}}
\renewcommand{\thetable}{A.\arabic{table}}

\section{Learning Details}

Our training framework is based on SERL\cite{luo2024serl}. Following SERL, we maintain both a demo buffer and a replay buffer for data storage. The demo buffer contains one pre-collected demonstration trajectory as well as transitions from automatic interventions during training. The replay buffer stores transitions collected during training together with transitions from automatic interventions. The hyperparameters used during training are summarized in Table~\ref{tab:hyperparameters}.

\section{Task Details}
We implement six tasks across three categories. The maximum episode length for all tasks is set to 300 time steps. For tasks implemented with the UR5 robot, the action scale is 0.005 m for translation and 0.05 rad for rotation. For tasks implemented with the Franka robot, the action scale is 0.01 m for translation and 0.06 rad for rotation. The translation ranges of all tasks relative to the initial pose are listed in Table~\ref{tab:translation_range}, and the rotation ranges are listed in Table~\ref{tab:rotation_range}.

\textit{The insertion tasks} demand high-precision control in constrained contact 
settings. In both USB insertion and plug insertion, the initial distance 
between the gripper and the target configuration is approximately 15~cm. 
Success is defined as complete insertion of the object into the corresponding 
port or socket.

\textit{The hanging tasks} require fine-grained control of both pose and trajectory to 
ensure correct geometric constraint interactions between the object's hole and 
the hook. The initial end-effector distance from the target hanging 
configuration is approximately 15~cm for the correction tape and spoon tasks, 
and approximately 20~cm for the hanger task. The task is considered successful 
when the hole of the correction tape or spoon, or the curved part of the 
hanger, establishes contact with the curved region of the hook.

\textit{The hinge-based task} consists of drawer pulling. The robot must first hook the 
drawer handle and then pull the drawer outward by 5~cm. The initial distance 
between the hook and the handle is approximately 10~cm. Success is achieved 
when the drawer is displaced by 5~cm. The main challenge lies in maintaining 
stable contact between the hook and the handle after initial engagement, while 
the drawer moves under the kinematic constraint of the hinge joint.

\begin{table}[!t]
\centering
\caption{Training hyperparameters.}
\begin{tabular}{lc}
\toprule
Parameter & Value \\
\midrule
Proprio Encoder Size & 64 \\
Policy MLP Size & 256 $\times$ 256 \\
Critic MLP Size & 256 $\times$ 256 \\
Discount Factor & 0.97 \\
Optimizer & Adam \\
Learning Rate & $3\times10^{-4}$ \\
Batch Size & 256 \\
\bottomrule
\end{tabular}
\label{tab:hyperparameters}
\end{table}

\begin{table}[!t]
\centering
\caption{Translation ranges relative to the initial pose for all tasks.}
\begin{tabular}{lccc}
\toprule
Task & X (m) & Y (m) & Z (m) \\
\midrule
USB Insertion & [-0.03, 0.03] & [-0.03, 0.03] & [-0.16, 0.01] \\
Plug Insertion & [-0.03, 0.03] & [-0.03, 0.03] & [-0.16, 0.01] \\
Correction Tape Suspension & [-0.04, 0.04] & [-0.04, 0.04] & [-0.16, 0.01] \\
Hanger Suspension & [-0.04, 0.04] & [-0.04, 0.04] & [-0.21, 0.01] \\
Spoon Suspension & [-0.04, 0.04] & [-0.04, 0.04] & [-0.15, 0.01] \\
Drawer Opening & [-0.01, 0.01] & [-0.035, 0.01] & [-0.1, 0.04] \\
\bottomrule
\end{tabular}
\label{tab:translation_range}
\end{table}

\begin{table}[!t]
\centering
\caption{Rotation ranges relative to the initial pose for all tasks.}
\begin{tabular}{lccc}
\toprule
Task & X (rad) & Y (rad) & Z (rad) \\
\midrule
USB Insertion & [-0.01, 0.01] & [-0.01, 0.01] & [-0.01, 0.01] \\
Plug Insertion & [-0.01, 0.01] & [-0.01, 0.01] & [-0.01, 0.01] \\
Correction Tape Suspension & [-0.0225, 0.0225] & [-0.0225, 0.0225] & [-0.0225, 0.0225] \\
Hanger Suspension & [-0.0225, 0.0225] & [-0.0225, 0.0225] & [-0.0225, 0.0225] \\
Spoon Suspension & [-0.0225, 0.0225] & [-0.0225, 0.0225] & [-0.0225, 0.0225] \\
Drawer Opening & [-0.0225, 0.0225] & [-0.0225, 0.0225] & [-0.0225, 0.0225] \\
\bottomrule
\end{tabular}
\label{tab:rotation_range}
\end{table}

\end{document}